\documentclass[letterpaper]{article} 
\usepackage{aaai25}  
\usepackage{times}  
\usepackage{helvet}  
\usepackage{courier}  
\usepackage[hyphens]{url}  
\usepackage{graphicx} 
\usepackage{natbib}  
\usepackage{caption} 
\usepackage{algorithm}
\usepackage{algorithmic}

\usepackage{amsmath}
\usepackage{amssymb}
\usepackage{amsthm}
\usepackage{subfigure}
\usepackage{multirow}
\newtheorem{theorem}{Theorem}

\usepackage{newfloat}
\usepackage{listings}
\DeclareCaptionStyle{ruled}{labelfont=normalfont,labelsep=colon,strut=off} 
\floatstyle{ruled}
\newfloat{listing}{tb}{lst}{}
\floatname{listing}{Listing}
\title{VKFPos: A Learning-Based Monocular Positioning with Variational Bayesian Extended Kalman Filter Integration}
\author {
    Jian-Yu Chen,\textsuperscript{\rm 1}
    Yi-Ru Chen,\textsuperscript{\rm 1}
    Yin-Qiao Chang\textsuperscript{\rm 1}, 
    Che-Ming Li\textsuperscript{\rm 2}, \\
    Jann-Long Chern\textsuperscript{\rm 3}, 
    Chih-Wei Huang\textsuperscript{\rm 1} 
}
\affiliations {
    \textsuperscript{\rm 1}Department of Communication Engineering, National Central University, Taoyuan, Taiwan\\
    \textsuperscript{\rm 2}AIoT BG - AI solution BU - SS R\&D Dept, ASUSTeK Computer Inc., Taipei, Taiwan\\
    \textsuperscript{\rm 3}Department of Mathematics, National Taiwan Normal University, Taipei, Taiwan\\
    \{111523039, 112523045, 110523036\}@cc.ncu.edu.tw, Ming\_Li@asus.com, \\
    chern@gapps.ntnu.edu.tw, cwhuang@ce.ncu.edu.tw
    
}

\begin{document}

\maketitle

\begin{abstract}
This paper addresses the challenges in learning-based monocular positioning by proposing VKFPos, a novel approach that integrates Absolute Pose Regression (APR) and Relative Pose Regression (RPR) via an Extended Kalman Filter (EKF) within a variational Bayesian inference framework. Our method shows that the essential posterior probability of the monocular positioning problem can be decomposed into APR and RPR components. This decomposition is embedded in the deep learning model by predicting covariances in both APR and RPR branches, allowing them to account for associated uncertainties. These covariances enhance the loss functions and facilitate EKF integration. Experimental evaluations on both indoor and outdoor datasets show that the single-shot APR branch achieves accuracy on par with state-of-the-art methods. Furthermore, for temporal positioning, where consecutive images allow for RPR and EKF integration, VKFPos outperforms temporal APR and model-based integration methods, achieving superior accuracy.
\end{abstract}

\section{Introduction}

Visual positioning techniques are increasingly crucial for the advancement of intelligent systems, including applications in augmented/mixed reality, smart factories, education, entertainment, and robotics~\cite{chen2017smart, li2019survey}.
In particular, the concept of monocular positioning~\cite{engel2014lsd, mur2015orb}, utilizing a single basic camera, is gradually emerging as an attractive alternative. 
Advantages like lightweight, cost-effectiveness, and minimal calibration make monocular positioning promising for future applications. However, significant challenges hinder widespread adoption~\cite{yang2018challenges, shu2021slam}, such as managing dynamic objects, adapting to varying illumination, and lacking depth information. These issues significantly impede practical use.

In recent years, machine learning approaches have gained attention as a solution to address the limitations of traditional methods in monocular positioning. 
The introduction of Absolute Pose Regression (APR) methods, pioneered by PoseNet~\cite{kendall2015posenet}, focuses on directly computing the absolute six degrees of freedom (6DoF) pose from a single image using convolutional neural networks (CNNs). 
Subsequent advances in PoseNet, such as PoseNet16~\cite{kendall2016modelling} and PoseNet17~\cite{kendall2017geometric}, named after the year they were published, extend the original work by considering uncertainty or adding learning weights to the loss function to enhance performance, respectively.
As these methods learn rotation in the form of 4-DoF quaternions, which are over-parameterized for rotation, the logarithm of a unit quaternion has been applied to represent the rotation in PoseNet+$\log q$~\cite{brahmbhatt2018geometry} for better representation of rotation.
Moreover, AtLoc~\cite{wang2020atloc} further improves precision by using an attention~\cite{vaswani2017attention} mechanism to focus on more geometrically robust features.
These methods, which solely utilize a single image to regress the absolute pose of the camera, are referred to as \emph{single-shot} positioning in the following content.

On the other hand, some approaches leverage multiple images to incorporate temporal information and strengthen the inter-image relationships, thereby enhancing performance. Examples of such methods include VidLoc~\cite{clark2017vidloc}, MapNet~\cite{brahmbhatt2018geometry}, and AtLoc+~\cite{wang2020atloc}, all of which belong to the category of \emph{temporal} positioning.
Methods like ms-Transformer~\cite{shavit2021learning} and DFNet~\cite{chen2022dfnet}, while leveraging advanced architectures such as transformers or direct feature matching, still struggle with overfitting and generalization across diverse scenes.
These techniques either utilize sequential images as input or design their loss functions to impose constraints on the distances between consecutive predictions. Despite these efforts to improve accuracy, pure APR methods still face certain limitations in terms of accuracy and robustness.

Given that camera motion tends to be continuous and smooth, LSTM-KF~\cite{coskun2017long} leverages long short-term memory (LSTM) to learn the current pose by considering all previous observations and states, or \cite{franccani2023transformer} leverages the Transformer network to understand the video stream, thereby considering trajectory information in the prediction process. 
However, relying solely on constant velocity or constant acceleration assumptions may not fully capture the dynamics present in the system, potentially leading to inaccuracies in estimation.
In contrast, Relative Pose Regression (RPR)~\cite{wang2017deepvo, li2018undeepvo} takes a different approach from APR, focusing on determining the relative motion between successive pairs of images.
RPR offers higher accuracy by directly estimating relative poses, but it faces challenges with accumulated errors when converting these to absolute poses due to its dependence on past temporal information. APR, on the other hand, provides drift-free predictions but tends to be less accurate. Hence, integrating both methods is essential to leverage their respective strengths and mitigate their weaknesses.

Integration methods can be divided into two main categories: model-based and optimization-based.
Model-based aims commonly employ recurrent neural networks (RNNs) to learn the integration strategies of RPR and APR predictions, such as ViPR~\cite{ott2020vipr}.
Another common strategy is to employ a shared visual encoder with the expectation that RPR can assist APR, as demonstrated in~\cite{valada2018deep}.
Despite the widespread use of RNN, it has become evident that the model-based integration approach lacks stability and does not consistently outperform optimization-based integration methods~\cite{ott2020vipr}.
This observation prompts the exploration of optimization-based approaches that may offer greater stability and improved performance compared to the reliance on RNNs.

The primary optimization-based methods for integrating predictions from multiple sensors include Pose Graph Optimization (PGO) and the Extended Kalman Filter (EKF). While PGO can achieve higher accuracy by optimizing a greater number of states, EKF is notable for its computational efficiency, making it well-suited for real-time applications where rapid processing is essential~\cite{laviola2003comparison}.

One notable work that has applied the EKF in learning-based positioning integration tasks is~\cite{zhou2020kfnet}. This study distinguishes itself by optimizing the predicted outcomes of the EKF, APR, and RPR simultaneously, considering posterior, likelihood, and prior probabilities concurrently. While this approach introduces a novel perspective, it deviates from the traditional Bayesian principles applied to the EKF. Such deviation may add complexity and raise concerns regarding its adherence to established Bayesian principles, potentially impacting the validity and reliability of the EKF framework.
Another significant work that integrates EKF principles is presented in~\cite{moreau2022coordinet}. This study aims to apply EKF to smooth trajectories and eliminate outliers effectively. However, it is important to note that the focus of this work is primarily on trajectory smoothing and outlier removal, without explicitly incorporating RPR in their methodology. The omission of RPR is a critical limitation given its essential role in capturing the relative motion between consecutive image pairs, which is fundamental for accurate positioning tasks.

In this study, we propose VKFPos, a novel lightweight monocular positioning approach that integrates APR and RPR using a variational Bayesian extended Kalman filter. The recursive structure of the Kalman filter leverages past state information to enhance current state prediction and trajectory forecasting. A key aspect of VKFPos is its innovative training paradigm for absolute and relative pose estimators, grounded in variational Bayesian inference.\\
The main contributions are as follows:
\begin{itemize}
    \item We introduce a robust and theoretically-founded method for integrating learned absolute and relative poses through an extended Kalman filter under the framework of variational Bayesian inference. We show that the essential posterior probability of the monocular positioning problem can be effectively decomposed into APR and RPR components.
    \item This decomposition is realized in the deep learning model by predicting the covariances in both APR and RPR branches and formulating them in the loss functions. This allows the branches to account for associated uncertainties, thereby improving the model's generalization capability.
    \item The results of performance evaluations on both indoor and outdoor datasets demonstrate that the single-shot APR branch achieves accuracy comparable to state-of-the-art methods. Moreover, in temporal positioning, where consecutive images enable RPR and EKF integration, VKFPos outperforms temporal APR and model-based integration methods, delivering superior accuracy\nolinebreak
    \footnote{The source code is avaliable at https://github.com/IPCLab/VKFPos}.
    
\end{itemize}


%
\section{Learning-Based Monocular Positioning with EKF Integration} \label{sec:arch}
%

\begin{figure}
    \centering
    \includegraphics[width=1\linewidth]{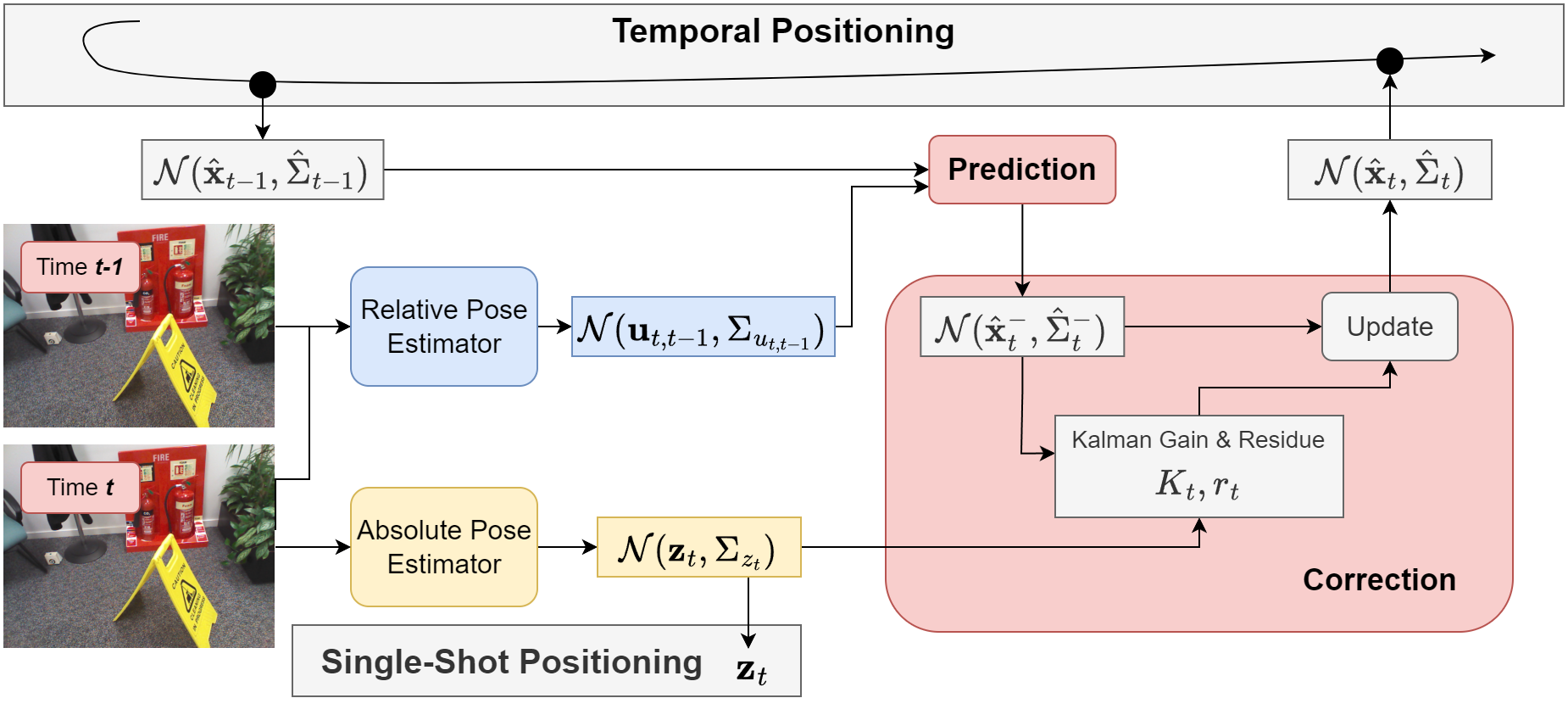}
    \caption{The architecture of learning-based monocular positioning with extended Kalman filter integration, VKFPos.}
    \label{fig:architecture}
\end{figure}

In this section, we present the proposed long-term probabilistic trajectory integration approach. The architecture is illustrated in Figure~\ref{fig:architecture}, where both the relative pose estimator and the absolute pose estimator predict their respective pose distributions and feed into an EKF to optimize the trajectory.

The relative pose estimator aims to predict the relative pose distribution, serving as the control input in the integration system. The absolute pose estimator predicts the absolute pose distribution, which can serve as the \emph{single-shot positioning} pose and is also utilized in the system correction stage as a measurement.
The EKF first performs a prediction step, utilizing the relative pose distribution to estimate the next state. This is followed by a correction step, in which the absolute pose distribution is used to refine the state estimate. The EKF computes the Kalman gain and residual, updating the state and covariance accordingly.
After the prediction and correction stages through the EKF, the system outputs a final pose that integrates the entire trajectory using both relative and absolute poses. This final pose is known as \emph{temporal positioning}.
Each component is further detailed in the following sections.

\subsection{Absolute Pose Estimator}

The absolute pose estimator is responsible for regressing the 6DoF global pose of the camera with its covariance.
This prediction is assumed to follow a normal distribution as 
\begin{equation}
    \mathbf{z}^{\text{truth}}_t \in \mathcal{N}(\mathbf{z}_t, \Sigma_{\mathbf{z}_t}),
\end{equation}
where $\mathbf{z}_t\in \mathbb{R}^6 \cong\mathfrak{s e}(3)$, containing both global translation $\mathbf{z}_x$ and rotation $\mathbf{z}_\theta$, and $\Sigma_{\mathbf{z}_t}\in \mathbb{R}^{6\times 6}$ are the pose predicted mean and covariance at time $t$, respectively.
The Lie algebra $\mathfrak{s e}(3)$ is the tangent space to the Lie group's manifold~\cite{sola2018micro}, enabling smooth interpolation and compact representation.
This allows us to model the covariance associated with each predicted pose, which is crucial for robust and reliable localization.
Here, $\mathbf{z}_t^{\text{truth}}\in \mathbb{R}^6 \cong\mathfrak{s e}(3)$ is the truth value of the pose, which is the provided ground truth in the datasets.

\begin{figure}
    \centering
    \includegraphics[width=1\linewidth]{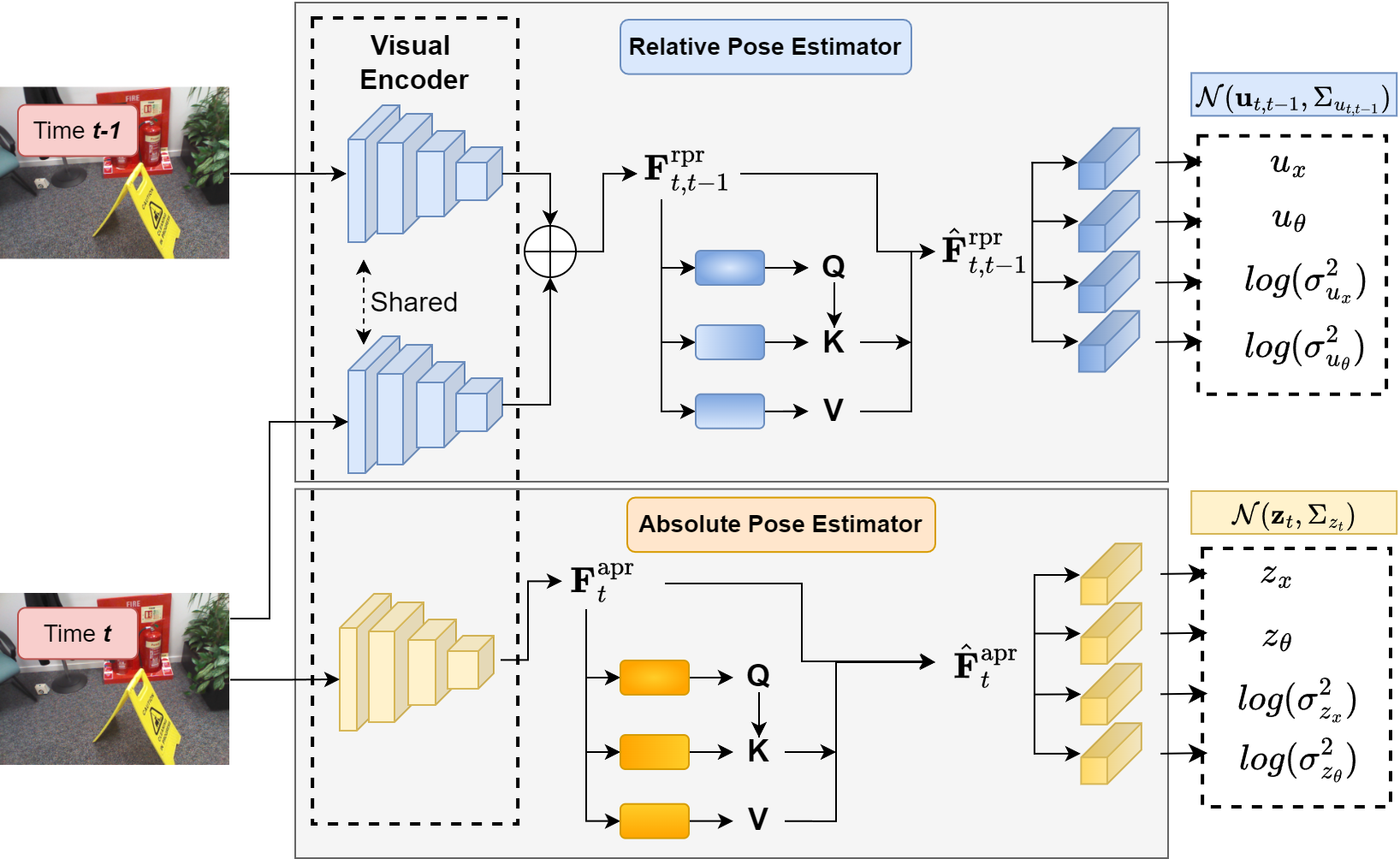}
    \caption{The upper branch is the relative pose estimator and the lower one is the absolute pose estimator, the visual encoder is ResNet34~\cite{he2016deep}}
    \label{fig:pose_estimator}
\end{figure}

As depicted in the lower part of Figure~\ref{fig:pose_estimator}, the absolute pose estimator is initiated from a visual encoder with a single inputted image at time $t$.
After a global average pooling layer reduces the feature dimension to $d_m$ denoted as $\mathbf{F}^\text{apr}_t$, a single non-local style self-attention module~\cite{vaswani2017attention} is employed to focus on spatial features as follows, 
\begin{equation}
    \hat{\mathbf{F}}^\text{apr}_t = \alpha(\text{Softmax}(\text{Q}\text{K}^T)\text{V}) + \mathbf{F}^\text{apr}_t, \label{eq:attention}
\end{equation}
where Q is a set of queries, K is a set of keys, V is a set of values, $\alpha$ is the scaled vectors, all of them are the output of a fully connected network with input $\mathbf{F}^\text{apr}_t$.
This architecture is inspired by AtLoc~\cite{wang2020atloc}, which is considered lightweight and accurate.

Differentiating from AtLoc, we further predict the covariance information in the final layer.
The output layer consists of four fully connected layers with input $\hat{\mathbf{F}}^\text{apr}_i$ responsible for predicting the mean values, $\mathbf{z_x}$ and $\mathbf{z_\theta}$ for absolute translation and rotation, as well as their respective covariance matrices, $\log(\Sigma_{\mathbf{z_x}})$ and $\log(\Sigma_{\mathbf{z_\theta}})$.
Notably, log covariance matrices are applied to ensure numerical stability~\cite{kendall2017uncertainties}.
The values will be recovered by taking the $\exp(.)$ operation after predicting to prevent values from reaching zero so that the covariance matrix defined as follows can ensure positive definite,
\begin{equation}
    \Sigma_{\mathbf{z}} = 
    \begin{bmatrix}
        \sigma_\mathbf{z_x}^2\mathbf{I}_3 &  \mathbf{0}_3 \\
        \mathbf{0}_3 & \sigma_\mathbf{z_\theta}^2\mathbf{I}_3
    \end{bmatrix}.
\end{equation}

We assert that the covariance matrix of each prediction should be diagonal, where the off-diagonal elements are zero, since each element in the predicted 6DoF pose is independent~\cite{moreau2022coordinet}. 
This independence implies that there are no relationships between two elements.


\subsection{Relative Pose Estimator}

The relative pose estimator is also tasked with regressing the 6DoF relative motion between consecutive images, along with its covariance matrix, assumed to follow a normal distribution as
\begin{equation}
    \mathbf{u}^{\text{truth}}_{t, t-1} \in \mathcal{N}(\mathbf{u}_{t, t-1}, \Sigma_{\mathbf{u}_{t, t-1}}),
\end{equation}
where $\mathbf{u}_{t, t-1}\in \mathbb{R}^6 \cong\mathfrak{se}(3)$ represents the combined relative translation and rotation between time steps $t-1$ and $t$, and $\Sigma_{u_{t, t-1}} \in \mathbb{R}^{6\times 6}$ denotes the covariance matrix associated with $\mathbf{u}_{t, t-1}$.
Additionally, $\mathbf{u}^{\text{truth}}_{t, t-1} \in \mathbb{R}^6 \cong\mathfrak{se}(3)$ represents the relative motion of the ground truth, calculated by $\mathbf{z}_{t-1}^{\text{truth}}$ and $\mathbf{z}_t^{\text{truth}}$.

As depicted in the upper part of Figure~\ref{fig:pose_estimator}, the relative pose estimator begins with a shared visual encoder that receives consecutive images as input, followed by concatenating their output features. 
Subsequently, a single non-local style self-attention module is applied to learn spatial and temporal information, as Eq.~\eqref{eq:attention}.
Through the attention network, the model can effectively prioritize feature changes associated with motion.

Similarly to the absolute pose estimator, four fully connected layers are employed to predict the mean of the relative motion, $\mathbf{u_x}$ and $\mathbf{u_\theta}$, representing relative translation and rotation, respectively, along with their respective log covariance matrices, $\log(\Sigma_\mathbf{u_x})$ and $\log(\Sigma_\mathbf{u_\theta})$. Upon applying the $\exp(.)$ operation, the covariance matrix of the relative pose is defined as

\begin{equation}
    \Sigma_{\mathbf{u}} = 
    \begin{bmatrix}
        \sigma_\mathbf{u_x}^2\mathbf{I}_3 &  \mathbf{0}_3 \\
        \mathbf{0}_3 & \sigma_\mathbf{u_\theta}^2\mathbf{I}_3
    \end{bmatrix},
\end{equation}
ensuring positive definiteness.

\subsection{Extended Kalman Filter Integration}\label{subsec:EKF_fusion}

The EKF is a recursive algorithm used to estimate the state of a dynamic system among uncertainty. 
To successfully apply the EKF, certain assumptions must be met. 
The first assumption, known as the Markov Chain Assumption, suggests that the current state is conditionally independent of all previous states, given the most recent state.
This assumption is inherently satisfied in the positioning task.
The second assumption, the Gaussian Noise Assumption, implies that the noise in the system, including both the process model and the measurement model, follows a Gaussian distribution. 
The previous statement that assumes that every state is Gaussian satisfies the second assumption.


As the EKF uses both estimates and knowledge of measurement distributions to find a distribution for the better estimate~\cite{charles2018kalman}, the EKF filter can be expressed as a Bayesian optimization.
Through the prior $\mathcal{N}(\hat{\mathbf{x}}^{-}_{t}, \hat{\Sigma}^{-}_{t})$ obtained from the prediction step and the likelihood $\mathcal{N}(\mathbf{z}_{t}, \Sigma_{z_t})$ from the correction step, we can get the optimal distribution $\mathcal{N}(\hat{\mathbf{x}}_{t}, \hat{\Sigma}_{t})$ according to following Bayes' theorem.

In the context of the EKF, the posterior distribution at time $t$ is conditioned on the previous state $\hat{\mathbf{x}}_{t-1}$, the control unit $\mathbf{u}_{t, t-1}$, and the measurement $\mathbf{z}_t$ can be expressed as $\,p(\hat{\mathbf{x}}_{t}| \hat{\mathbf{x}}_{t-1}, \mathbf{u}_{t, t-1}, \mathbf{z}_t)$, which satisfies the Markov property.
This formulation captures the recursive nature of the EKF, where each posterior distribution serves as the prior for the subsequent time step, continuously refining the state estimate based on new measurements and control inputs.

\begin{theorem}
    If $\hat{\mathbf{x}}_{t}$ is only conditioned to the measurement $\mathbf{z}_t$, the posterior distribution of $\hat{\mathbf{x}}_{t}$ given $\hat{\mathbf{x}}_{t-1}$, $\mathbf{u}_{t, t-1}$, and $\mathbf{z}_t$ will be proportional to the product of the measurement likelihood and the transition probability, expressed as follows:
    \begin{gather}
    p(\hat{\mathbf{x}}_{t}| \hat{\mathbf{x}}_{t-1}, \mathbf{u}_{t, t-1}, \mathbf{z}_t) \nonumber \\
    \propto 
    p(\mathbf{z}_t|\hat{\mathbf{x}}_{t})
    p(\mathbf{u}_{t, t-1}|\hat{\mathbf{x}}_{t}, \hat{\mathbf{x}}_{t-1}).\label{eq:finalform}
\end{gather}
\end{theorem}

\begin{proof}
\begin{gather}
    p(\hat{\mathbf{x}}_{t}| \hat{\mathbf{x}}_{t-1}, \mathbf{u}_{t, t-1}, \mathbf{z}_t) \label{eq:posterior} \\   
    =
    \frac{p(\mathbf{z}_t|\mathbf{u}_{t, t-1}, \hat{\mathbf{x}}_{t-1}, \hat{\mathbf{x}}_{t}) p(\hat{\mathbf{x}}_{t}, \hat{\mathbf{x}}_{t-1}, \mathbf{u}_{t, t-1})}{p(\mathbf{z}_t)}\label{eq:likelihoodandprior}.
\end{gather}

According to the Bayes' theorem, the Eq.~\eqref{eq:posterior} can be factored into Eq.~\eqref{eq:likelihoodandprior}, which consists of the measurement likelihood $p(\mathbf{z}_t|\mathbf{u}_{t, t-1}, \hat{\mathbf{x}}_{t-1}, \hat{\mathbf{x}}_{t})$ and the transition probability $p(\hat{\mathbf{x}}_{t}, \hat{\mathbf{x}}_{t-1}, \mathbf{u}_{t, t-1})$ over prior $p(\mathbf{z}_t)$.
This equation highlights the importance of integrating both the measurement information and the prior state estimate to obtain an updated posterior distribution.

The subsequent step is to simplify the calculation by leveraging the conditional independence assumptions inherent in the EKF.
Thus, the measurement likelihood can be simplified as follows,
\begin{equation}
    p(\mathbf{z}_t|\mathbf{u}_{t, t-1}, \hat{\mathbf{x}}_{t-1}, \hat{\mathbf{x}}_{t}) = p(\mathbf{z}_t|\hat{\mathbf{x}}_{t}), 
\end{equation}
which indicates that the measurement is solely dependent on the current state and is independent of control input and the previous state.

Similarly, for the prior distribution, the simplification can be expressed as,
\begin{align}
    p(\hat{\mathbf{x}}_{t}, \hat{\mathbf{x}}_{t-1}, \mathbf{u}_{t, t-1})
    &=
    p(\mathbf{u}_{t, t-1}| \hat{\mathbf{x}}_{t-1}, \hat{\mathbf{x}}_{t}) p(\hat{\mathbf{x}}_{t-1}, \hat{\mathbf{x}}_{t}) \nonumber \\
    &\propto 
    p(\mathbf{u}_{t, t-1}|\hat{\mathbf{x}}_{t}, \hat{\mathbf{x}}_{t-1}),
\end{align}
which shows the prior is proportional to the control unit according to the current and previous state.

Thus the Eq.~\eqref{eq:likelihoodandprior} can be further simplified as follows,
\begin{gather}
    p(\hat{\mathbf{x}}_{t}| \hat{\mathbf{x}}_{t-1}, \mathbf{u}_{t, t-1}, \mathbf{z}_t) \nonumber \\
    = \frac{p(\mathbf{z}_t|\mathbf{u}_{t, t-1}, \hat{\mathbf{x}}_{t-1}, \hat{\mathbf{x}}_{t}) p(\mathbf{u}_{t, t-1}| \hat{\mathbf{x}}_{t-1}, \hat{\mathbf{x}}_{t}) p(\hat{\mathbf{x}}_{t-1}, \hat{\mathbf{x}}_{t})}{p(\mathbf{z}_t)} \nonumber \\
    \propto 
    p(\mathbf{z}_t|\hat{\mathbf{x}}_{t})
    p(\mathbf{u}_{t, t-1}|\hat{\mathbf{x}}_{t}, \hat{\mathbf{x}}_{t-1}).
\end{gather}
\end{proof}
This final form demonstrates how the EKF integrates information from both the measurement and the state transition to update the posterior distribution of the state estimate. Specifically, in our scheme, the measurement corresponds to the APR, while the transition model reflects the RPR. Thus, the posterior distribution captures the combined influence of the current APR measurement and the dynamics represented by the RPR transition model.

\section{Training the EKF Integrated Model}\label{sec:training}

\subsection{APR and RPR Loss Functions}

Benefiting from the inherent flexibility of Bayes' theorem in EKF, we partition the training process into relative and absolute positioning branches.
According to Eq.~\eqref{eq:finalform}, we can effectively optimize both the prior and likelihood components independently.
This enables us to formulate the maximized function more effectively.

However, the likelihood $p(\mathbf{z}_t|\hat{\mathbf{x}}_{t})$ and the transition probability $p(\mathbf{u}_{t, t-1}|\hat{\mathbf{x}}_{t}, \hat{\mathbf{x}}_{t-1})$ still can not be solved analytically.
To address this, we aim to approximate them using variational inference with a simpler parameterized distribution $q(\mathbf{z}_t|\mathbf{I}_{t}; \theta_{\text{APR}})$ and $q(\mathbf{u}_{t, t-1}|\mathbf{I}_{t}, \mathbf{I}_{t-1}; \theta_{\text{RPR}})$, where $\theta_{\text{APR}}$ and $\theta_{\text{RPR}}$ denote the absolute and relative model parameters, respectively.
We optimize these parameters by maximizing the evidence lower bound as follows,
\begin{align}
    \hat{\theta}_{\text{APR}} &=  \mathop{\arg\max}\limits_{\mathbf{\theta}} \prod_{t} q(\mathbf{z}_t|\mathbf{I}_{t}; \theta_{\text{APR}}),
    \label{eq:optimalapr}\\
    \hat{\theta}_{\text{RPR}} &= \mathop{\arg\max}\limits_{\mathbf{\theta}} \prod_{t} q(\mathbf{u}_{t, t-1}|\mathbf{I}_{t}, \mathbf{I}_{t-1}; \theta_{\text{RPR}}), \label{eq:optimalrpr}
\end{align}
these density functions serve as the surrogate objective function that balances the trade-off between the complexity of the approximation and the fidelity to the true distribution.

To learn the parameters $\theta_\text{APR}$ and $\theta_{\text{RPR}}$, we take the negative logarithm of Eq.~\eqref{eq:optimalapr} and Eq.~\eqref{eq:optimalrpr} as our loss function to minimize and define as follows,
\begin{align}
    \mathcal{L}_{\text{RPR}} &= \frac{1}{2}\frac{\|\mathbf{u}-\mathbf{u}^\text{truth}\|^2_2}{\Sigma_u} + \frac{1}{2} \log(\Sigma_u),\label{eq:loss_rpr} \\
    \mathcal{L}_{\text{APR}} &= \frac{1}{2}\frac{\|\mathbf{z}-\mathbf{z}^\text{truth}\|^2_2}{\Sigma_z} + \frac{1}{2} \log(\Sigma_z).\label{eq:loss_apr}
\end{align}
Unlike other approaches that utilize geometric loss~\cite{kendall2017geometric} to balance the weight between translation and rotation, our model employs this loss function to learn the true distribution of each timestamp, providing a more accurate depiction of the camera's state. 
By separately minimizing these two loss functions, Eq.~\eqref{eq:loss_rpr} and Eq.~\eqref{eq:loss_apr}, and integrated with EKF, the posterior of the whole system can be maximized, leading to the optimal system.


\subsection{EKF Updating and Correction}

\begin{figure}
    \centering
    \includegraphics[width=1\linewidth]{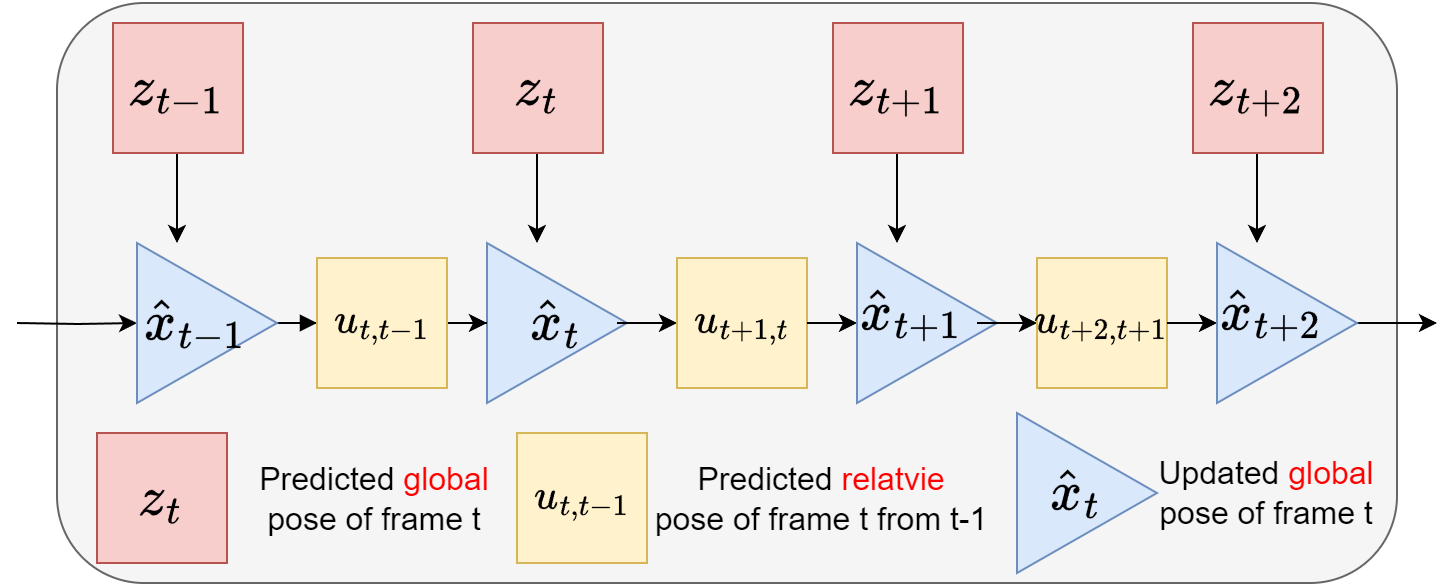}
    \caption{This is a visualization of how EKF does in the entire scheme. Absolute poses act as measurements, relative poses act as the control unit, thus the $\hat{\mathbf{x}}_t$ is the final prediction that integrates both information by EKF.}
    \label{fig:pose graph}
\end{figure}

Then, we demonstrate how we leverage EKF to enhance the robustness of our predictions.
Figure~\ref{fig:architecture} provides a visual representation of how the EKF is seamlessly integrated into our framework, and Figure~\ref{fig:pose graph} shows the pose graph, which highlights the operation of the EKF within the overall scheme.

In the EKF algorithm, the process model and the measurement model are defined as follows,
\begin{align}
    \hat{\mathbf{x}}_t &= f(\hat{\mathbf{x}}_{t-1}, \mathbf{u}_{t, t-1}),\label{eq:transition} + \mathcal{N}(0, \Sigma_{\mathbf{\mathbf{u}_{t, t-1}}})\\
   \mathbf{z}_t &= h(\hat{\mathbf{x}}_t) +\mathcal{N}(0, \Sigma_{\mathbf{z_t}}),
\end{align}
where $f(.)$ is the state transition function for combining the previous state and the relative pose to the new state.
$h(.)$ is the measurement function to measure the current state, which is exactly the identity in our approach, and $\mathcal{N}(0, \Sigma_{\mathbf{u}_{t, t-1}})$ and $\mathcal{N}(0, \Sigma_{\mathbf{z_t}})$ are the white noise, respectively, in the process model and the measurement model.

And for the prediction step, the process involves using the predicted relative pose $\mathbf{u}_{t, t-1}$ to estimate the next state $\hat{\mathbf{x}}_t^-$ as Eq.~\eqref{eq:transition_x} and predicting the next covariance $\hat{\Sigma}_t^-$ from $\Sigma_{\mathbf{u}_{t, t-1}}$ as Eq.~\eqref{eq:transition_cov}.
\begin{align}
    \mathbf{x}_t^- &=  \hat{\mathbf{x}}_{t-1} \oplus \mathbf{u}_{t, t-1}\label{eq:transition_x} ,\\
    \hat{\Sigma}_t^- &= F_t \hat{\Sigma}_{t-1} F_t^T + \Sigma_{\mathbf{u}_{t, t-1}}\label{eq:transition_cov},\\
    F_t &:= \frac{\partial f}{\partial \mathbf{x}} \bigg|_{\hat{\mathbf{x}}_{t-1}, \mathbf{u}_{t, t-1}},
\end{align}
where $\oplus$ is the manifold update operation, which is needed since the states are under $\mathfrak{s e}(3)$.
Here, $F_t$ is the Jacobian matrix of $f(.)$ at time $t$.

Moving to the correction step, the measurement residual $\mathbf{r}_t$ and the Kalman gain $\mathbf{K}_t$ are sequentially computed through measurement and the state from the prediction step as follows
\begin{align}
    \mathbf{r}_t &= \mathbf{z}_t \ominus h(\hat{\mathbf{x}}_t^-) \\
    \mathbf{K}_t &= \hat{\Sigma}_t^- H^T (H_t \hat{\Sigma}_t^- H_t^T + \Sigma_{\mathbf{z}_t})^{-1}, \\
    H_t &:= \frac{\partial h}{\partial \mathbf{x}} \bigg|_{\hat{\mathbf{x}}_{t}^-},
\end{align}
where $\ominus$ is also the manifold operation representing the inverse operation of $\oplus$.
Here $H_t$ is the Jacobian matrix of $h(.)$.
Therefore, the corrected state estimate $\hat{\mathbf{x}}_t$ and the updated state covariance $\hat\Sigma_t$ is obtained as follows,
\begin{align}
    \hat{\mathbf{x}}_t &= \hat{\mathbf{x}}_t^- \oplus \mathbf{K}_t \mathbf{r}_t, \\
    \hat\Sigma_t &= (\mathbf{I} - \mathbf{K}_t H_t) \hat\Sigma_t^-.
\end{align}
And finally, the prediction distribution $\mathcal{N}(\hat{\mathbf{x}}_t, \hat{\Sigma}_t)$ is well defined and produced.

\section{Experimental Results}\label{sec:experiment}

\subsection{Implementation Details}

We input pairs of consecutive images into our model, concurrently training both the absolute and relative branches. 
To maintain consistency, we resize the shorter side length of the input images to $256$ pixels and utilize a ResNet34 backbone pretrained on ImageNet for the visual encoder. 
The images undergo normalization to ensure a zero mean and a standard deviation of $1$. 
We employ a batch size of $64$ and set the learning rate to $5 \times 10^{-5}$ for both branches. 
A dropout rate of $0.5$ is applied to both branches to enhance generalization, and the Adam optimizer~\cite{kingma2014adam} is used for optimization. 
These hyperparameters are determined through grid search, with the learning rate ranging from $10^{-4}$ to $10^{-6}$, the batch size varying from $4$ to $64$.
Additionally, we incorporate a weight decay of $5 \times 10^{-4}$ to further mitigate overfitting. 

When training on the Oxford RobotCar dataset, we further apply random ColorJitter augmentation, setting values of $0.7$ for brightness, contrast, and saturation, and $0.5$ for hue. 
This augmentation step is observed to be crucial for improving generalization across various weather and time conditions~\cite{wang2020atloc}. 
Note that we do not apply random cropping during training, as done in AtLoc, and we also avoid center cropping during inference. 
This decision is made to maintain consistency with the corresponding ground truth poses, as random cropping could potentially alter the spatial context of the images.


The scheme is implemented using PyTorch~\cite{NEURIPS2019_9015}, a widely adopted deep learning framework known for its flexibility, ease of use, and extensive community support.
All experiments are conducted on a personal computer equipped with an Intel Core i9-12900K CPU @ 3.2GHz $\times$ 16 and an Nvidia GeForce RTX 3090Ti GPU.

\begin{table*}
\resizebox{2\columnwidth}{!} {%
\begin{tabular}{c|lllllllll}
\multicolumn{1}{l|}{} & Method & Chess & Fire & Heads & Office & Pumpkin & Kitchen & Stairs & Average \\ \hline
\multirow{6}{*}{\rotatebox{90}{Single-shot}} & PoseNet15 & 0.32m, 8.12° & 0.47m, 14.4° & 0.29m, 12.0° & 0.48m, 7.68° & 0.47m, 8.42° & 0.59m, 8.64° & 0.47m, 13.8° & 0.44m, 10.4° \\
 & PoseNet16 & 0.37m, 7.24° & 0.43m, 13.7° & 0.31m, 12.0° & 0.48m, 8.04° & 0.61m, 7.08° & 0.58m, 7.54° & 0.48m, 13.1° & 0.47m, 9.81° \\
 & PoseNet17 & 0.14m, 4.50° & 0.27m, 11.8° & 0.18m, 12.10° & 0.20m, 5.77° & 0.25m, 4.82° & 0.24m, 5.52° & 0.37m, 10.60° & 0.24m, 7.87° \\
 & PoseNet+log q & 0.11m, 4.29° & 0.27m, 12.13° & 0.19m, 12.15° & 0.19m, 6.35° & 0.22m, 5.05° & 0.25m, \pmb{5.27°} & 0.30m, 11.29° & 0.22m, 8.07° \\
 & AtLoc & \pmb{0.10m}, \pmb{4.07°} & \pmb{0.25m}, \pmb{11.4°} & \pmb{0.16m}, \pmb{11.8°} & \pmb{0.17m}, \pmb{5.34°} & \pmb{0.21m}, 4.37° & 0.23m, 5.42° & \pmb{0.26m}, \pmb{10.5°} & \pmb{0.20m}, \pmb{7.56°} \\
 & VKFPos(Single-Shot) & \pmb{0.10m}, 4.80° & 0.27m, 12.95° & 0.18m, 12.26° & 0.20m, 7.10° & 0.22m, \pmb{4.14°} & \pmb{0.22m}, 5.30° & 0.28m, 10.9° & \pmb{0.20m}, 8.20° \\ \hline \hline
\multicolumn{1}{l|}{\multirow{4}{*}{\rotatebox{90}{Temporal}}} & VidLoc & 0.18m, NA & 0.26m, NA & 0.14m, NA & 0.26m, NA & 0.36m, NA & 0.31m, NA & 0.26m, NA & 0.25m, NA \\
\multicolumn{1}{l|}{} & MapNet & 0.08m, 3.25° & 0.27m, 11.69° & 0.18m, 13.25° & \pmb{0.17m}, \pmb{5.15°} & 0.22m, 4.02° & 0.23m, 4.93° & 0.30m, 12.08° & 0.21m, 7.77° \\
\multicolumn{1}{l|}{} & AtLoc+ & 0.10m, \pmb{3.18°} & \pmb{0.26m}, 10.8° & \pmb{0.14m}, 11.4° & \pmb{0.17m}, 5.16° & 0.20m, 3.94° & 0.16m, 4.90° & 0.29m, 10.2° & 0.19m, 7.08° \\
\multicolumn{1}{l|}{} & VKFPos(Ours) & \pmb{0.09m}, 3.26° & 0.27m, \pmb{10.2°} & \pmb{0.14m}, \pmb{10.07°} & 0.18m, 6.06° & \pmb{0.19m}, \pmb{3.84°} & \pmb{0.15m}, \pmb{4.32°} & \pmb{0.24m}, \pmb{9.52°} & \pmb{0.18m}, \pmb{6.75°}
\end{tabular}}
\caption{Positioning results on the indoor 7-Scenes dataset. The median errors in translation (cm) and rotation (°) are reported, with the best results highlighted.}
\label{tab:7scenes}
\end{table*}

\subsection{Baselines}

To evaluate the effectiveness of our model, we conducted experiments on both indoor and outdoor datasets, specifically the 7-Scenes dataset~\cite{shotton2013scene} and the Oxford RobotCar dataset~\cite{maddern20171robotcar}. 
Each dataset presents unique challenges that our model must address.

Given our use of ResNet as the visual encoder backbone, we conduct comparisons with algorithms that employ a similar architecture. 
To ensure credibility, we exclusively select papers with available codes, enabling a fair and relevant benchmarking process. 
Baseline algorithms are categorized into \emph{single-shot} methods, which consider only single images, and \emph{temporal} methods, which utilize sequential images.

For the 7Scenes dataset, \emph{single-shot} methods include PoseNet15~\cite{kendall2015posenet}, PoseNet16~\cite{kendall2016modelling}, PoseNet17~\cite{kendall2017geometric}, PoseNet+$\log q$~\cite{brahmbhatt2018geometry}, AtLoc~\cite{wang2020atloc}, and our single-shot positioning of VKFPos. 
For the Oxford RobotCar dataset, PoseNet+$\log q$~\cite{brahmbhatt2018geometry}, and AtLoc are chosen for comparisons, as other methods lack results on this dataset.
Notably, the single-shot positioning of VKFPos shares similarities with AtLoc but differs in loss design and covariance prediction, potentially enhancing generalization ability.

\emph{Temporal} methods, which utilize sequential images, include VidLoc~\cite{clark2017vidloc}, MapNet, AtLoc+~\cite{wang2020atloc}, and VKFPos for the 7Scenes dataset. 
These methods leverage the temporal relationships between consecutive frames to improve the accuracy and stability of pose prediction. 
For the Oxford RobotCar dataset, comparisons are limited to MapNet and AtLoc+\cite{wang2020atloc}, as VidLoc\cite{clark2017vidloc} lacks numerical results.

Additionally, VKFPos should be compared with other model-based integration approaches that utilize RNN-based networks to refine predictions or directly enhance performance. Examples include LSTM-KF~\cite{coskun2017long}, and ViPR~\cite{ott2020vipr}. However, these works have not released their experimental code for validation. Therefore, we compare with them separately.

\subsection{Performance on 7-Scenes Indoor Dataset}

The upper part of Table~\ref{tab:7scenes} presents the results of single-shot methods employing ResNet34 as the visual encoder backbone. 
In particular, PoseNet15~\cite{kendall2015posenet} deviates from this configuration; however, its significance in machine learning-based approaches should be acknowledged.
In this comparison, our focus is on evaluating the performance of VKFPos in single-shot positioning for the sake of fairness.

The results demonstrate that single-shot positioning of VKFPos outperforms most approaches and achieves performance levels comparable to the state-of-the-art single-shot APR method, AtLoc~\cite{wang2020atloc}. 
It is important to note that our model is trained with an emphasis on covariance estimation and might need a more distinguished dataset, thus only attending a similar performance with AtLoc in the small indoor dataset.
For example, compared to other methods, VKFPos achieves the best performance on the larger dataset (Pumpkin and Kitchen) in rotation and translation, respectively.
These two scenes have the most sequences compared to other scenes, which provides more information and special features to learn.

The lower part of Table~\ref{tab:7scenes} presents a comparison of temporal methods. 
VKFPos exhibits superior translation errors compared to VidLoc, which focuses solely on translation in its approach.
Additionally, our approach demonstrates outstanding performance with MapNet and AtLoc+, which also leverage sequential images as input for positioning. 
An important observation is the significant accuracy achieved by VKFPos, particularly in reducing translation and rotation errors, following the integration of relative motion information. 
This integration results in a notable improvement of up to $10\%$ in translation accuracy and $17.6\%$ in rotation accuracy, underscoring the effectiveness of our approach.

\begin{table}
\resizebox{\columnwidth}{!}
{%
\begin{tabular}{l|ccl}
Scene   & \multicolumn{1}{c}{LSTM-KF} & \multicolumn{1}{c}{ViPR} & \multicolumn{1}{c}{VKFPos}         \\ \hline
Chess   & 0.33m,   6.9°               & 0.22m,   7.89°           & \pmb{0.09m}, \pmb{3.26°}  \\
Fire    & 0.41m, 15.7°                & 0.38m, 12.74°            & \pmb{0.27m}, \pmb{10.2°}  \\
Heads   & 0.28m, 13.01°               & 0.21m, 16.41°            & \pmb{0.14m}, \pmb{10.07°} \\
Offices & 0.43m, 7.65°                & 0.35m, 9.59°             & \pmb{0.18m}, \pmb{6.06°}  \\
Pumpkin & 0.49m, 10.63°               & 0.37m, 8.45°             & \pmb{0.19m}, \pmb{3.84°}  \\
Kitchen & 0.57m, 8.53°                & 0.40m, 9.32°             & \pmb{0.15m}, \pmb{4.32°}  \\
Stairs  & 0.46m, 14.56°               & 0.31m, 12.65°            & \pmb{0.24m}, \pmb{9.52°}  \\
Average & 0.424m, 11.00°              & 0.32m, 11.01°            & \pmb{0.18m}, \pmb{6.75°} 
\end{tabular}%
}\caption{Comparison of results on the 7-Scenes benchmark for model-based integration approaches and our proposed optimization-based VKFPos. Median errors in translation and rotation are reported, with the best results highlighted.}
\label{tab:7scenes_hybrid}
\end{table}

Table~\ref{tab:7scenes_hybrid} presents a comparison of our optimization-based integration approach, VKFPos, with other model-based integration approaches on the 7-Scenes benchmark.
VKFPos demonstrates superior performance across all scenes compared to those model-based methods, such as LSTM-KF~\cite{coskun2017long} and ViPR~\cite{ott2020vipr}, and achieves significantly lower median errors in both translation and rotation in every scene. 
This remarkable improvement can be attributed to the effective consideration of absolute and relative pose information through the EKF, leveraging the strengths of both estimations.

The success of our approach lies in its adept integration of domain knowledge, both effectively and theoretically. 
By carefully optimizing each branch to capitalize on inherent assumptions in positioning tasks, we achieve more precise and robust pose estimations. 
This underscores the importance of leveraging domain-specific information in developing integration strategies, which demonstrates superiority over model-based integration approaches.

\begin{table}
\resizebox{\columnwidth}{!}{%
\begin{tabular}{l|llccc}
& Methods &  & LOOP & FULL & AVERAGE \\ \hline
\multicolumn{1}{c|}{\multirow{6}{*}{\rotatebox{90}{Single-shot}}} & \multirow{2}{*}{\begin{tabular}[c]{@{}l@{}}PoserNet+$\log q$\end{tabular}} & Mean & 25.29m, 17.45° & 125.6m, 27.10° & 75.45m, 22.27° \\
\multicolumn{1}{c|}{} &  & Median & 6.88m, 2.06° & 107.6m, 22.5° & 57.24m, 12.28° \\
\multicolumn{1}{c|}{} & \multirow{2}{*}{AtLoc} & Mean & 8.61m, 4.58° & 29.6m, 12.4° & 19.11m, 8.49° \\
\multicolumn{1}{c|}{} &  & Median & 5.68m, 2.23° & 11.1m, 5.28° & 8.39m, 3.76° \\
\multicolumn{1}{c|}{} & \multirow{2}{*}{\begin{tabular}[c]{@{}l@{}}VKFPos\\ (Single-Shot)\end{tabular}} & Mean & \pmb{6.9m}, \pmb{3.84°} & \pmb{13.25m}, \pmb{2.81°} & \pmb{10.08m}, \pmb{3.33°} \\
\multicolumn{1}{c|}{} &  & Median & \pmb{5.61m}, \pmb{1.83}° & \pmb{10.82m}, \pmb{1.67°} & \pmb{8.22m}, \pmb{1.75°} \\ \hline \hline
\multirow{6}{*}{\rotatebox{90}{Temporal}} & \multirow{2}{*}{MapNet} & Mean & \multicolumn{1}{l}{8.76m, 3.46°} & \multicolumn{1}{l}{41.4m, 12.5°} & 25.08m, 7.98° \\
 &  & Median & \multicolumn{1}{c}{5.79m, 1.54°} & \multicolumn{1}{l}{17.94m, 6.68°} & 11.87m, 4.11° \\
 & \multirow{2}{*}{AtLoc+} & Mean & \multicolumn{1}{l}{7.82m, 3.62°} & \multicolumn{1}{l}{21.0m, 6.15°} & 14.41m, 4.89° \\
 &  & Median & \multicolumn{1}{c}{4.34m, 1.92°} & \multicolumn{1}{l}{\pmb{6.40m}, 1.50°} & \pmb{5.37m}, 1.71° \\
 & \multirow{2}{*}{VKFPos(Ours)} & Mean & \multicolumn{1}{l}{\pmb{4.7m}, \pmb{2.41°}} & \multicolumn{1}{l}{\pmb{9.77m}, \pmb{2.69°}} & \pmb{7.24m}, \pmb{2.55°} \\
 &  & Median & \multicolumn{1}{c}{\pmb{4.66m}, \pmb{1.73°}} & \multicolumn{1}{l}{8.13m, \pmb{1.32°}} & 6.4m, \pmb{1.53°}
\end{tabular}}
\caption{Positioning results on the outdoor RobotCar dataset. Median and mean errors in translation(cm) and rotation(°) are reported, with the best results highlighted.}
\label{tab:robotcar}
\end{table}

\subsection{Performance on RobotCar Outdoor Dataset}

\begin{figure}[htb]
    \centering
    \begin{tabular}{cc}
     \includegraphics[width=0.23\textwidth]{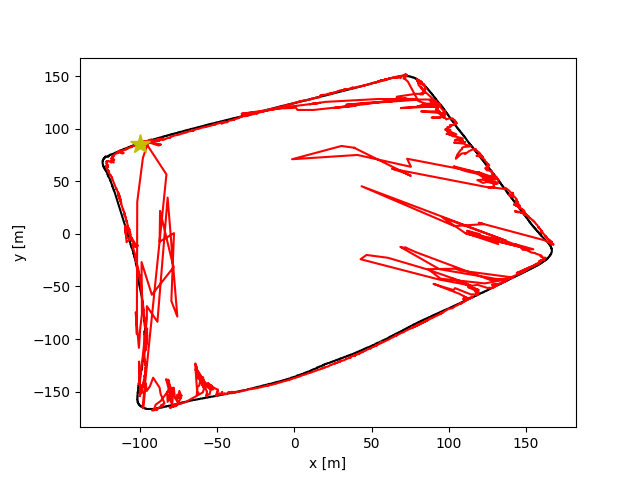}&
     \includegraphics[width=0.23\textwidth]{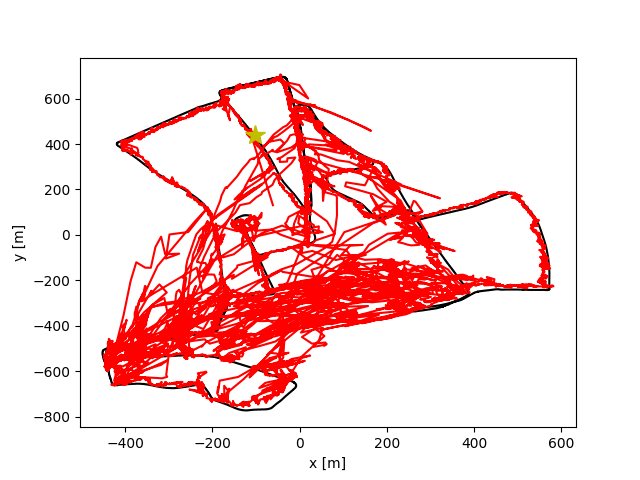}\\
     \includegraphics[width=0.23\textwidth]{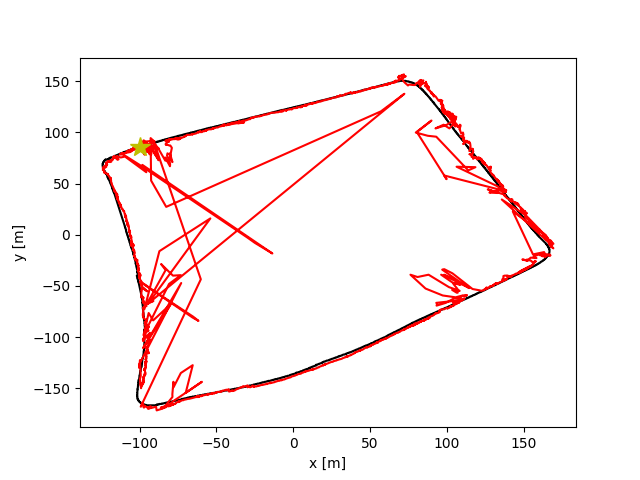}&
     \includegraphics[width=0.23\textwidth]{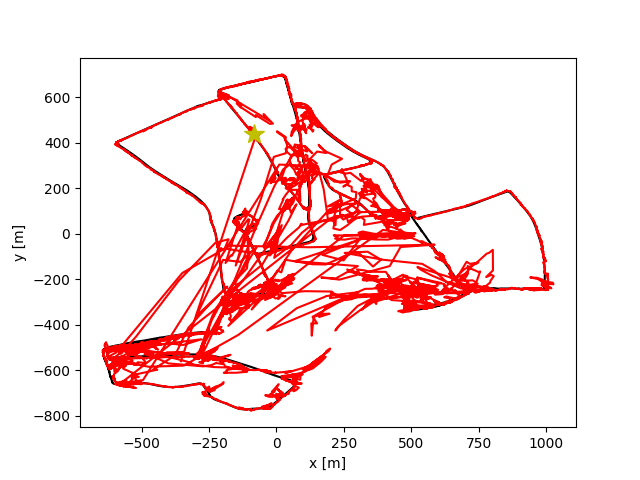}\\
     \includegraphics[width=0.23\textwidth]{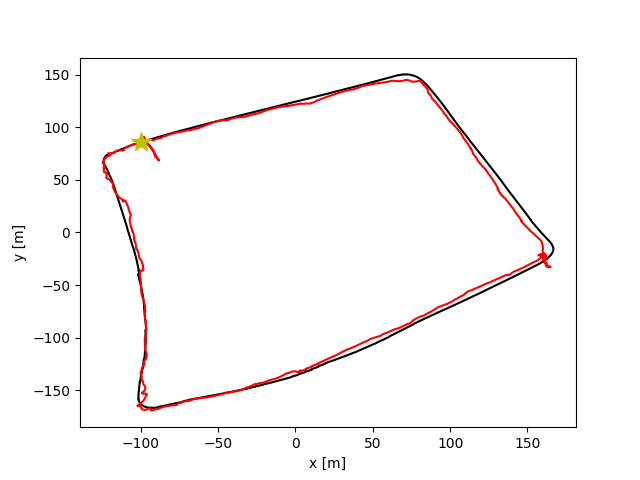}&
     \includegraphics[width=0.23\textwidth]{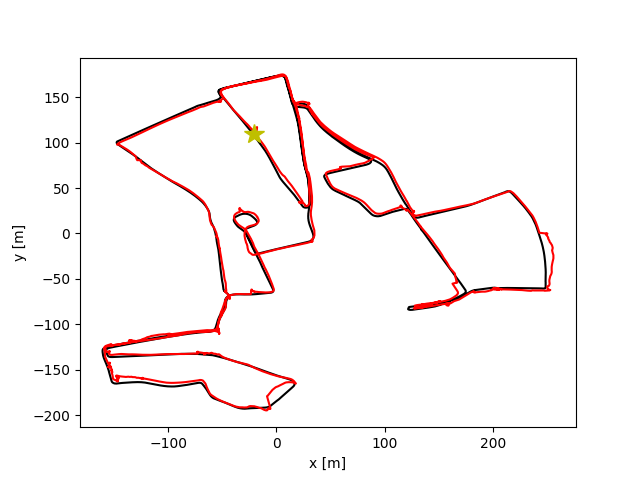}\\
     \multicolumn{1}{c}{LOOP} & \multicolumn{1}{c}{FULL}\\
     \end{tabular}
\caption{Temporal positioning trajectory of \textbf{MapNet(Upper), AtLot+(Center), VKFPos(Lower)} on Oxford RobotCar Dataset. The ground truth is shown in black lines and the red lines are the prediction, while the start represents the starting point.}
\label{fig:robotcar_traj_ekf}
\end{figure}

The upper part of Table~\ref{tab:robotcar} provides a comprehensive comparison of VKFPos with existing single-shot methods on the large outdoor dataset. 
The single-shot positioning of VKFPos consistently outperforms PoseNet+$\log q$ and AtLoc in most of the sequences in terms of mean and median translation and rotation errors. 
Specifically, VKFPos achieves notable improvements in mean errors, demonstrating approximately $47.2\%$ and $60.7\%$ improvement in mean translation error and mean rotation error, respectively, indicating enhanced robustness and accuracy in pose estimation. 
Although the median translation error of VKFPos is comparable to that of AtLoc, VKFPos exhibits significantly greater stability and robustness, particularly in handling outliers. This highlights its effectiveness in diverse and challenging real-world scenarios. Importantly, these results underscore the efficacy of our proposed covariance learning approach in enhancing generalization capability in single-shot APR tasks. Consequently, VKFPos stands out as a promising solution for large outdoor datasets.

The lower part of Table~\ref{tab:robotcar} presents a comparative analysis of our proposed VKFPos with existing temporal methods, VKFPos demonstrates competitive performance across all sequences in terms of mean and median errors in translation and rotation. 
Particularly noteworthy is our method's ability to consistently achieve lower mean errors compared to MapNet and AtLoc+, gaining approximately $49.7\%$ and $47.8\%$ improvement over AtLoc+.
In sequences such as \emph{LOOP} and \emph{FULL}, VKFPos exhibits significant reductions in both mean translation and rotation errors, indicating its superior accuracy and robustness in handling complex real-world environments. 

In particular, VKFPos achieves only the mean translation error $9.77$ m and the mean rotation error $2.69^{\circ}$ in the entire \emph{FULL} sequence with a long route of $9562$ m, which shows stability and can be observed in Fig.~\ref{fig:robotcar_traj_ekf}.
These results demonstrate the effectiveness of our proposed method in leveraging temporal information to enhance pose estimation accuracy, making it a promising solution for applications in autonomous navigation and robotics.

\section{Conclusion}\label{sec:conclusion}

In conclusion, we present a novel integration approach for learning-based monocular positioning through the EKF, effectively leveraging both APR and RPR to address their inherent limitations. Our method is grounded in variational Bayesian inference, providing a solid theoretical foundation. The integration of EKF significantly improves the robustness and accuracy in positioning. Experimental results demonstrate our approach's effectiveness, achieving competitive single-shot accuracy and surpassing existing methods in temporal performance on the 7-Scenes and Oxford RobotCar datasets. The consistent and reliable covariance estimates further validate VKFPos as a valuable advancement in learning-based monocular positioning.



\bibliography{VKFPos_arXiv}

\end{document}